\documentclass{article}

\usepackage{microtype}
\usepackage{graphicx}
\usepackage{subfigure}
\usepackage{booktabs} 

\usepackage{hyperref}

\usepackage[accepted]{icml2025}

\usepackage{amsmath}
\usepackage{amssymb}
\usepackage{mathtools}
\usepackage{amsthm}

\usepackage{booktabs} 
\usepackage{multirow} 
\usepackage{array}    

\usepackage[capitalize,noabbrev]{cleveref}

\theoremstyle{plain}

\theoremstyle{definition}

\theoremstyle{remark}

\usepackage[textsize=tiny]{todonotes}

\def\b{\ensuremath\boldsymbol}

\icmltitlerunning{On the Relation of State Space Models and Hidden Markov Models}

\begin{document}

\twocolumn[
\icmltitle{On the Relation of State Space Models and Hidden Markov Models}



\icmlsetsymbol{equal}{*}

\begin{icmlauthorlist}
\icmlauthor{Aydin Ghojogh}{equal,aaa}
\icmlauthor{M.Hadi Sepanj}{equal,bbb}
\icmlauthor{Benyamin Ghojogh}{equal,bbb}
\end{icmlauthorlist}

\icmlaffiliation{aaa}{Thunder Bay, ON, Canada}
\icmlaffiliation{bbb}{Waterloo, ON, Canada}

\icmlcorrespondingauthor{Aydin Ghojogh}{aghojogh@lakeheadu.ca}
\icmlcorrespondingauthor{M.Hadi Sepanj}{mhsepanj@uwaterloo.ca}
\icmlcorrespondingauthor{Benyamin Ghojogh}{bghojogh@uwaterloo.ca}

\icmlkeywords{Machine Learning, ICML}

\vskip 0.3in
]



\printAffiliationsAndNotice{\icmlEqualContribution} 

\begin{abstract}
State Space Models (SSMs) and Hidden Markov Models (HMMs) are foundational frameworks for modeling sequential data with latent variables and are widely used in signal processing, control theory, and machine learning. Despite their shared temporal structure, they differ fundamentally in the nature of their latent states, probabilistic assumptions, inference procedures, and training paradigms. Recently, deterministic state space models have re-emerged in natural language processing through architectures such as S4 and Mamba, raising new questions about the relationship between classical probabilistic SSMs, HMMs, and modern neural sequence models.

In this paper, we present a unified and systematic comparison of HMMs, linear Gaussian state space models, Kalman filtering, and contemporary NLP state space models. We analyze their formulations through the lens of probabilistic graphical models, examine their inference algorithms—including forward–backward inference and Kalman filtering—and contrast their learning procedures via Expectation–Maximization and gradient-based optimization. By highlighting both structural similarities and semantic differences, we clarify when these models are equivalent, when they fundamentally diverge, and how modern NLP SSMs relate to classical probabilistic models. Our analysis bridges perspectives from control theory, probabilistic modeling, and modern deep learning.
\end{abstract}


\section{Introduction}

Sequential data arises naturally in a wide range of domains, including speech recognition, Natural Language Processing (NLP), control systems, finance, and biological signal analysis. A central challenge in modeling such data is capturing temporal dependencies through latent representations that summarize past information while remaining computationally tractable. Two of the most influential frameworks for addressing this challenge are Hidden Markov Models (HMMs) and State Space Models (SSMs), both of which introduce latent state variables to explain observed temporal structure.

HMMs represent latent dynamics using discrete-valued states that evolve according to a Markov chain, with observations generated conditionally on the current state. This formulation has proven effective in applications where latent modes are naturally categorical, such as speech recognition \cite{juang1991hidden,mark2024application} and biological sequence analysis \cite{krogh1998introduction,yoon2009hidden}. In contrast, classical SSMs, originating in control theory \cite{ogata2010modern}, model latent states as continuous-valued vectors evolving according to linear or nonlinear dynamical equations, often corrupted by Gaussian noise. The Linear Gaussian SSM (LG-SSM), together with Kalman filtering and smoothing \cite{kalman1960new,kalman1963mathematical}, provides a mathematically elegant and computationally optimal solution for inference in such systems.

Although HMMs and SSMs share a common temporal dependency structure, their modeling assumptions lead to fundamentally different interpretations of uncertainty, inference mechanisms, and learning algorithms. HMMs rely on discrete probabilistic inference via the forward–backward algorithm \cite{baum1966statistical,baum1970maximization}, while linear Gaussian SSMs employ Kalman filtering and smoothing \cite{kalman1960new,kalman1963mathematical} to propagate continuous Gaussian beliefs over latent trajectories. Parameter estimation in both frameworks is commonly performed using the Expectation–Maximization (EM) algorithm, yet the quantities computed in the E-step and the updates in the M-step differ substantially due to the nature of the latent variables.

Recently, state space models have gained renewed attention in NLP through architectures such as S4 \cite{gu2022efficiently} and Mamba \cite{gu2024mamba}. These models discretize continuous-time linear systems to obtain efficient, long-range sequence models, but depart from classical SSMs by removing stochasticity and adopting deterministic state transitions trained by backpropagation \cite{rumelhart1986learning,ghojogh2024backpropagation}. As a result, modern NLP SSMs blur the traditional boundaries between probabilistic graphical models, recurrent neural networks, and dynamical systems, leading to confusion regarding their relationship to HMMs and linear Gaussian SSMs. 

Probabilistic state space modeling has attracted increasing attention in areas such as image processing \cite{roesser2003discrete,nazemi2024particle}, tracking \cite{sepanj2025uncertainty}, etc.
Moreover, some recent papers have focused on theoretical analysis of state space models, e.g., the relation of SSM and attention mechanism has been analyzed in \cite{ghodsi2025many}. Likewise, this paper is an analytical paper discussing the relation of SSM and HMM.
The goal of this paper is to clarify these relationships. We provide a unified exposition of HMMs, linear Gaussian SSMs, Kalman filtering, and modern NLP state space models from a probabilistic graphical modeling perspective. We compare their formulations, inference procedures, and learning algorithms, with particular emphasis on the role of Expectation–Maximization and the interpretation of latent uncertainty. By explicitly contrasting probabilistic and deterministic state space formulations, we aim to demystify the connections between classical models and modern sequence architectures, and to provide a conceptual bridge between control theory, probabilistic modeling, and contemporary deep learning.

The remainder of this paper is organized as follows. Section \ref{section_ssm_hmm} introduces the mathematical formulations of HMMs and SSMs, with particular emphasis on linear Gaussian SSMs and their relationship to Kalman filtering. Section \ref{section_EM} discusses inference and learning in these models, including forward--backward inference for HMMs and Kalman filtering and smoothing for SSMs, as well as parameter estimation via the Expectation--Maximization algorithm. Section \ref{section_comparison} compares SSMs and HMMs in terms of their formulations, probabilistic graphical models, inference, learning, and uncertainty interpretation. Finally, Section \ref{section_conclusion} summarizes the key insights and concludes the paper with more organized comparisons. 

\section{Introduction to State Space Models and Hidden Markov Models}\label{section_ssm_hmm}

\subsection{Continuous-Time State Space Model}

\textit{State Space Models (SSMs)} have their roots in control theory \cite{ogata2010modern}, where they serve as a formal framework for modeling dynamical systems and representing their internal states \cite{birkhoff1927dynamical}. An SSM establishes a linear relationship between an input signal $\b{x}(t)$ and an output signal $\b{y}(t)$ through a latent state variable $\b{h}(t)$. The model is composed of two fundamental equations:
\begin{enumerate}
\item State equation: Governs the temporal evolution of the system’s hidden state as determined by its current value and the external input:
\begin{align}\label{equation_continuous_SSM_state_equation}
\b{h}'(t) = \b{A}\b{h}(t) + \b{B}\b{x}(t),
\end{align}
where $\b{x}(t)$ denotes the input signal, $\b{y}(t)$ represents the output signal, $\b{h}(t)$ corresponds to the latent state, and $\b{h}'(t)$ denotes the derivative of $\b{h}(t)$ with respect to time.
\item Output equation: Specifies how the observable output is computed from the latent state and the input:
\begin{align}\label{equation_continuous_SSM_output_equation}
\b{y}(t) = \b{C}\b{h}(t) + \b{D}\b{x}(t).
\end{align}
\end{enumerate}
The matrices $\b{A}$, $\b{B}$, $\b{C}$, and $\b{D}$ constitute the learnable parameters of the model. Among them, $\b{A}$ is known as the state transition matrix, $\b{B}$ as the input mapping matrix, and $\b{C}$ as the observation matrix.

Conceptually, the state equation in Eq. (\ref{equation_continuous_SSM_state_equation}) expresses the rate of change of the latent state as a superposition of linear effects arising from the current state and the input signal. The output equation in Eq. (\ref{equation_continuous_SSM_output_equation}) links the hidden representation and the input to the observed output. 
The solution of this coupled system of differential equations yields the latent trajectory $\b{h}(t)$ and output $\b{y}(t)$, given the system matrices $\b{A}$, $\b{B}$, $\b{C}$, and $\b{D}$.

\subsection{Discrete-Time State Space Model}

The formulations in Eqs. (\ref{equation_continuous_SSM_state_equation}) and (\ref{equation_continuous_SSM_output_equation}) are expressed in continuous time. These continuous-time differential equations are impractical for discrete-time sequential data,
such as language, which is processed at discrete time steps.
To address this limitation, the continuous-time differential equations can be transformed into a discrete-time form, yielding the following update rules \cite{gu2022efficiently,gu2024mamba}:
\begin{align}
& \b{h}_{k+1} = \b{A} \b{h}_{k} + \b{B} \b{x}_k, \label{equation_discrete_SSM_state_equation} \\
& \b{y}_k = \b{C} \b{h}_k, \label{equation_discrete_SSM_output_equation}
\end{align}
where $\b{h}_k \in \mathbb{R}^s$ denotes the latent state at discrete time step $k$, $\b{x}_k \in \mathbb{R}^d$ represents the input at time step $k$, and $\b{y}_k \in \mathbb{R}^p$ is the corresponding output. The matrices $\b{A} \in \mathbb{R}^{s \times s}$, $\b{B} \in \mathbb{R}^{s \times d}$, and $\b{C} \in \mathbb{R}^{p \times s}$ are discrete-time counterparts of the continuous-time matrices $\b{A}$, $\b{B}$, and $\b{C}$ in Eqs. (\ref{equation_continuous_SSM_state_equation}) and (\ref{equation_continuous_SSM_output_equation}), respectively.
This formulation of SSM is used in Natural Language Processing (NLP), so we refer to it as \textit{NLP SSM} in this paper. 

By contrasting Eq. (\ref{equation_continuous_SSM_output_equation}) with Eq. (\ref{equation_discrete_SSM_output_equation}), it can be observed that the term $\b{D}\b{x}_k$ is omitted in the discrete formulation, which is equivalent to setting $\b{D} = \b{0}$. This simplification is adopted for computational efficiency, since the matrix $\b{C}$ is sufficient to account for the influence of the input $\b{x}_k$:
\begin{align*}
\b{y}_k &\overset{(\ref{equation_discrete_SSM_output_equation})}{=} \b{C} \b{h}_k
\overset{(\ref{equation_discrete_SSM_state_equation})}{=}
\b{C}(\b{A} \b{h}_{k-1} + \b{B} \b{x}_{k-1})
\\
&= \b{C}\b{A} \b{h}_{k-1} + \b{C}\b{B} \b{x}_{k-1},
\end{align*}
demonstrating that $\b{C}$ implicitly acts on the input $\b{x}$ through the state update.

\subsection{Linear Gaussian State Space Model and Kalman Filter}

The Eqs. (\ref{equation_discrete_SSM_state_equation}) and (\ref{equation_discrete_SSM_output_equation}) are deterministic SSM. It is possible to include randomness to have a stochastic SSM. The \textit{Linear Gaussian SSM (LG-SSM)}, which is basically a linear dynamical system with Gaussian noise \cite{kalman1963mathematical}, is formulated as \cite{kitagawa1996linear,elvira2022graphical}:
\begin{align}
& \b{h}_{k+1} = \b{A} \b{h}_{k} + \b{B} \b{x}_k + \b{w}_k, \quad &&\b{w}_k \sim \mathcal{N}(\b{0}, \b{Q}), \label{equation_linear_gaussian_ssm_1} \\
& \b{y}_k = \b{C} \b{h}_k + \b{v}_k, \quad &&\b{v}_k \sim \mathcal{N}(\b{0}, \b{R}), \label{equation_linear_gaussian_ssm_2}
\end{align}
where the hidden state $\b{h}$ and the output $\b{y}$ are random variables with Gaussian distributions. 
The LG-SSM admits optimal inference via the \textit{Kalman filter} \cite{kalman1960new,kalman1963mathematical}, which provides recursive state estimation for this model. 
In this paper, we distinguish between probabilistic SSMs (specifically linear Gaussian SSMs) and deterministic NLP SSMs, both of which are commonly referred to as state space models in the literature.

\subsection{Hidden Markov Model}

\textit{Hidden Markov Model (HMM)} is a graphical model in which the latent states are categorical, belonging to a discrete set of states \cite{ghojogh2019hidden}:
\begin{align}
\b{h}_k \in \{1, \dots, K\}.
\end{align}
The formulation of HMM is as follows \cite{rabiner1989tutorial,cappe2005inference}: 
\begin{align}
\b{h}_k &\sim \text{Categorical}(\b{A}_{k-1,:}), \label{equation_HMM_1} \\
\b{y}_k &\sim \mathbb{P}(\b{y}_k \mid \b{h}_k), \label{equation_HMM_2}
\end{align}
where $\mathbb{P}(.)$ denotes the probability density distribution and $\b{A}$ is the stochastic state transition matrix. 
These equations can be stated as probability distributions:
\begin{align}
& \mathbb{P}(\b{h}_k \mid \b{h}_{k-1}) = \b{A}_{k-1,k}, \\
& \mathbb{P}(\b{y}_k \mid \b{h}_{k}) = \mathbb{P}(\b{y}_k \mid \b{h}_k = i).
\end{align}

\section{Inference and Learning in SSM and HMM}\label{section_EM}

\subsection{Joint Likelihood Distribution}

In both LG-SSM and HMM, the joint likelihood of the latent states and the outputs is:
\begin{equation}\label{equation_SSM_joint_distirbution}
\begin{aligned}
&p(\b{h}_{1:T}, \b{y}_{1:T})
\\
&~~~~~=
p(\b{h}_1)
\prod_{k=2}^T
p(\b{h}_k \mid \b{h}_{k-1}, \b{x}_{k-1})
\prod_{k=1}^T
p(\b{y}_k \mid \b{h}_k),
\end{aligned}
\end{equation}
for a sequence of $\{1, \dots, T\}$.
The joint log-likelihood in both LG-SSM and HMM is:
\begin{equation}\label{equation_SSM_joint_log_distirbution}
\begin{aligned}
&\log p(\b{h}_{1:T}, \b{y}_{1:T}) =
\log p(\b{h}_1) \\
&~~~~+ \sum_{k=2}^T
\log p(\b{h}_k \mid \b{h}_{k-1}, \b{x}_{k-1})
+ \sum_{k=1}^T
\log p(\b{y}_k \mid \b{h}_k).
\end{aligned}
\end{equation}

\subsection{Expectation--Maximization in HMM}\label{section_EM_HMM}

\subsubsection{E-Step in HMM}

The E-step in HMM computes the state occupancy probability $\gamma_k(i)$ and the state transition probability $\xi_k(i,j)$, using the forward--backward propagation:
\begin{align}
&\gamma_k(i) = \mathbb{P}(\b{h}_k = i \mid \b{y}_{1:T}), \label{equation_HMM_E_step_1} \\
&\xi_k(i,j) = \mathbb{P}(\b{h}_{k-1}=i, \b{h}_k = j \mid \b{y}_{1:T}). \label{equation_HMM_E_step_2}
\end{align}

\subsubsection{M-Step in HMM}

The M-step for HMM calculates the elements of the state transition matrix: 
\begin{align}
\b{A}_{i,j} = \frac{\sum_{k=2}^T \xi_k(i,j)}{\sum_{k=2}^T \gamma_{k-1}(i)},
\end{align}
which is interpreted as the expected transition counts or expected state counts. 

For detailed mathematics of expectation--maximization in HMM, refer to \cite{ghojogh2019hidden}.

\subsection{Expectation--Maximization in SSM}\label{section_EM_SSM}

\subsubsection{E-Step in SSM}

The E-step in SSM calculates the first moment $\mathbb{E}[\b{h}_k]$ and second moments $\mathbb{E}[\b{h}_k \b{h}_k^\top]$ and $\mathbb{E}[\b{h}_k \b{h}_{k-1}^\top]$, by Kalman smoothing \cite{kalman1960new}. 


\subsubsection{M-Step in SSM}

The M-step for HMM calculates the state transition matrix, also called the dynamics matrix:
\begin{align}
\b{A} = \Big( \sum_{k=2}^T \mathbb{E}[\b{h}_k \b{h}_{k-1}^\top] \Big) \Big( \sum_{k=2}^T \mathbb{E}[\b{h}_{k-1} \b{h}_{k-1}^\top] \Big)^{-1},
\end{align}
which is the least-squares regression using expected states.
Similar closed-form updates exist for the observation matrix $\b{C}$ and noise covariances, $\b{Q}$ and $\b{R}$, and are omitted here for brevity.

For detailed mathematics of expectation--maximization in SSM and Kalman smoothing, refer to \cite{simon2006optimal,sarkka2023bayesian}.

\section{Comparison of Hidden Markov Models and State Space Models}\label{section_comparison}




\subsection{Model Formulation}

Hidden Markov Models (HMMs) and State Space Models (SSMs) are latent-variable models for sequential data that differ primarily in the nature of their hidden states and transition dynamics. In an HMM, the latent state $\b{h}_k$ is a discrete random variable evolving according to a Markov chain, as in Eqs. (\ref{equation_HMM_1}) and (\ref{equation_HMM_2}).
This formulation models dynamics as probabilistic switching between a finite number of states.

In contrast, classical SSMs define continuous-valued latent states $\b{h}_k \in \mathbb{R}^d$ that evolve according to stochastic dynamical equations (\ref{equation_linear_gaussian_ssm_1}) and (\ref{equation_linear_gaussian_ssm_2}).
Here, uncertainty is expressed through continuous probability distributions over latent trajectories.

Modern NLP state space models, such as S4 \cite{gu2022efficiently} and Mamba \cite{gu2024mamba}, further modify this formulation by removing stochasticity and defining deterministic state updates obtained by discretizing continuous-time linear systems, as in Eqs. (\ref{equation_discrete_SSM_state_equation}) and (\ref{equation_discrete_SSM_output_equation}).
Unlike HMMs and classical SSMs, these models are deterministic and are not trained as probabilistic generative models. Although, a deterministic computation graph can still be embedded into a probabilistic model.

\subsection{Probabilistic Graphical Model Structure}

\begin{figure}[!t]
    \centering
    \includegraphics[width=1\linewidth]{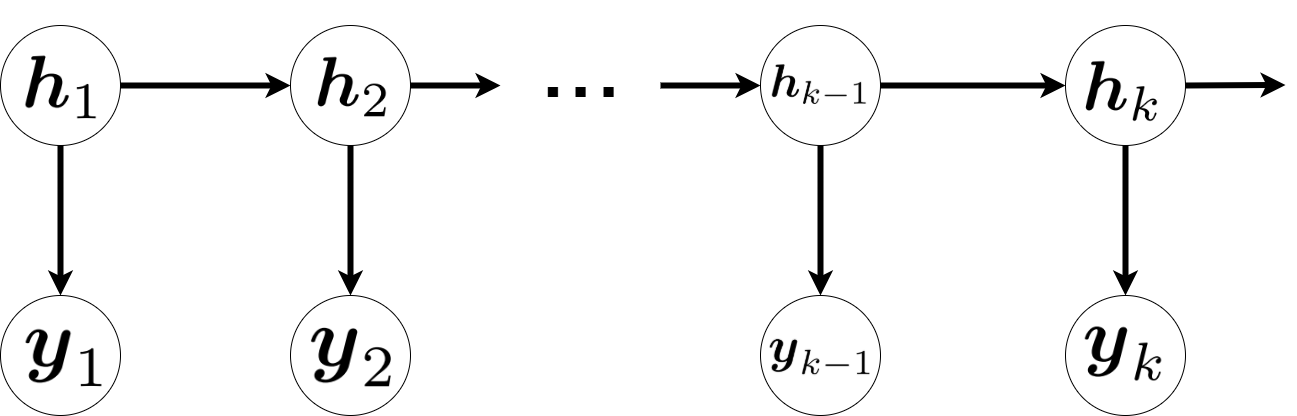}
    \caption{The Probabilistic Graphical Model (PGM) of the hidden Markov models and state space models. Note that state space models can also have input signal $\b{x}_k$ in its graphical model, but it is not included in this figure for the sake of easier comparison of hidden Markov models and state space models.}
    \label{figure_pgm}
\end{figure}

From a Probabilistic Graphical Model (PGM) perspective \cite{koller2009probabilistic}, HMMs and SSMs share the same temporal structure: a first-order Markov chain over latent variables, with observations conditionally independent given the latent state. The PGM of SSMs and HMMs is illustrated in Fig. \ref{figure_pgm}. The joint distribution factorizes as in Eq. (\ref{equation_SSM_joint_distirbution}), where $\b{h}_k$ is discrete in HMMs and continuous in SSMs.









Despite this identical graphical skeleton, the semantics of the nodes differ. In HMMs, latent variables correspond to categorical random variables, while in SSMs they correspond to continuous random vectors with Gaussian transitions. NLP SSMs retain the same computational dependency structure but drop probabilistic semantics altogether, yielding computational graphs rather than probabilistic graphical models.

\subsection{Inference by Forward--Backward or Kalman Filtering}

Inference in HMMs is performed using the forward--backward algorithm \cite{baum1966statistical,baum1970maximization,bishop2006pattern}, which implements sum--product message passing on the chain-structured PGM \cite{kschischang2002factor,ghojogh2019hidden}. The resulting posterior quantities, Eqs. (\ref{equation_HMM_E_step_1}) and (\ref{equation_HMM_E_step_2}), represent expected state occupancies and transitions.

In LG-SSMs, inference is carried out via Kalman filtering and smoothing \cite{kalman1960new}, which propagate Gaussian messages characterized by posterior means $\mathbb{E}[\b{h}_k \mid \b{y}_{1:T}]$ and covariances $\mathbb{E}[\b{h}_k\b{h}_k^\top \mid \b{y}_{1:T}]$.
The Kalman filter thus serves as the continuous-state analogue of the forward pass in HMM inference.

NLP SSMs require no probabilistic inference, as latent states are deterministic functions of inputs. In this case, inference reduces to a forward scan identical to that of a recurrent neural network.

\subsection{Learning and Expectation--Maximization}

\begin{table*}[t]
\centering
\begin{tabular}{l|cccc}
\hline
 & HMM & LG-SSM & Kalman Filter & NLP SSM (S4/Mamba) \\
\hline
Latent state & Discrete & Continuous & Continuous & Continuous \\
Time & Discrete & Discrete & Discrete & Discrete \\
Stochastic & Yes & Yes & Yes & No \\
PGM defined & Yes & Yes & N/A & No \\
Inference & Forward--Backward & Kalman smoother & Kalman filter & Forward scan \\
Training & EM & EM & N/A & Backpropagation \\
Uncertainty & State identity & Trajectory & Trajectory & None \\
Role & Probabilistic model & Probabilistic model & Algorithm & Deterministic model \\
\hline
\end{tabular}
\caption{Comparison of HMMs, SSMs, Kalman filtering, and modern NLP SSMs.}\label{table_comparison}
\end{table*}

Parameter learning in HMMs and SSMs is commonly performed using the Expectation--Maximization (EM) algorithm. As stated in Section \ref{section_EM_HMM}, in HMMs, the E-step consists of computing posterior state probabilities using forward--backward inference, while the M-step updates parameters via normalized counting and maximum-likelihood estimation.

As mentioned in Section \ref{section_EM_SSM}, in LG-SSMs, the E-step is implemented via Kalman smoothing, yielding posterior means and covariances of latent states. The M-step updates transition and observation matrices using least-squares estimation and covariance matching. The Kalman filter itself does not perform learning, but provides the inference machinery required for the E-step.

In contrast, NLP SSMs are trained discriminatively using gradient-based optimization, as the absence of latent stochasticity makes the EM algorithm unnecessary.

\subsection{Uncertainty Interpretation}

HMMs represent uncertainty over discrete latent states, emphasizing ambiguity in state identity. SSMs represent uncertainty over continuous latent trajectories, emphasizing uncertainty in system dynamics. NLP SSMs, by construction, do not represent uncertainty at the latent-state level and instead model deterministic dynamics optimized for predictive performance.

\section{Conclusion}\label{section_conclusion}

From a unifying perspective, HMMs, linear Gaussian SSMs, and modern NLP state space models can be viewed as occupying different points along a spectrum of latent-variable modeling. HMMs employ discrete latent states with probabilistic transitions, SSMs generalize this structure to continuous latent dynamics with Gaussian uncertainty, and NLP SSMs further specialize the framework by removing stochasticity entirely and optimizing deterministic state updates via gradient-based learning. While all three share a similar temporal dependency structure, their modeling assumptions lead to fundamentally different interpretations, inference procedures, and training paradigms.

Table \ref{table_comparison} summarizes comparisons and key distinctions among HMMs, Linear Gaussian State Space Models (LG-SSMs), the Kalman filter, and modern NLP state space models such as S4 and Mamba. The \textbf{latent state} row indicates whether the hidden variables are discrete (categorical) as in HMMs, or continuous as in SSMs and NLP SSMs. All models assume \textbf{discrete time} updates; this is purely the indexing of sequential steps and does not imply discreteness of the latent state. The \textbf{stochastic} row highlights whether the model incorporates probabilistic uncertainty in the latent dynamics: HMMs and LG-SSMs are stochastic, while NLP SSMs define deterministic transitions. The \textbf{PGM defined} row indicates whether the model can be interpreted as a classical Probabilistic Graphical Model (PGM); both HMMs and LG-SSMs have well-defined PGMs, while NLP SSMs do not due to their deterministic nature\footnote{More accurately, NLP SSMs as used in practice (e.g., S4 and Mamba) are deterministic and are not trained as probabilistic generative models. However, a deterministic computation graph can still be embedded into a probabilistic model.}. The Kalman filtering is an algorithm and not a model; hence, it does not admit a PGM. 

The \textbf{inference} row specifies the standard method for computing posterior information about latent states given observations. HMMs rely on the forward--backward algorithm to compute discrete posterior distributions over latent states, LG-SSMs use Kalman smoothing to obtain posterior means and covariances, the Kalman filter performs recursive filtering over continuous latent states, and NLP SSMs require only a forward deterministic pass. The \textbf{training} row distinguishes between training methods: HMMs and LG-SSMs are typically trained using the Expectation--Maximization (EM) algorithm, the Kalman filter itself does not perform learning but provides inference for EM in SSMs, and NLP SSMs are trained end-to-end using backpropagation \cite{rumelhart1986learning,ghojogh2024backpropagation}. 

The \textbf{uncertainty} row captures the interpretation of the latent state: HMMs quantify uncertainty over discrete state identity, LG-SSMs and the Kalman filter quantify uncertainty over continuous trajectories, and NLP SSMs do not represent latent uncertainty. 
Finally, the \textbf{role} row distinguishes between model classes and inference algorithms. HMM is a probabilistic generative model, i.e., a fully defined instance with discrete states and transitions. LG-SSM defines a probabilistic generative model, i.e., a continuous latent state model with Gaussian noise. The Kalman filter is not a model but it is an algorithm for inference within LG-SSMs. NLP SSMs are deterministic models---and not probabilistic models---which are optimized for sequence prediction. 
While probabilistic interpretations may be introduced after training, NLP SSMs are not specified or optimized as probabilistic generative models.

The Table \ref{table_comparison}, together with the detailed discussion above, emphasizes how identical graphical structures can yield fundamentally different models and inference procedures depending on the nature of the latent states and their probabilistic assumptions.

\bibliography{references}

@book{ogata2010modern,
  title     = {Modern Control Engineering},
  author    = {Ogata, Katsuhiko},
  edition   = {5},
  year      = {2010},
  publisher = {Prentice Hall}
}

@book{birkhoff1927dynamical,
  title={Dynamical systems},
  author={Birkhoff, George David},
  volume={9},
  year={1927},
  publisher={American Mathematical Soc.}
}

@inproceedings{gu2022efficiently,
  title={Efficiently Modeling Long Sequences with Structured State Spaces},
  author={Gu, Albert and Goel, Karan and Re, Christopher},
  booktitle={International Conference on Learning Representations},
  year={2022}
}

@inproceedings{gu2024mamba,
  title={Mamba: Linear-Time Sequence Modeling with Selective State Spaces},
  author={Gu, Albert and Dao, Tri},
  booktitle={First Conference on Language Modeling},
  year={2024}
}

@article{kalman1963mathematical,
  title={Mathematical description of linear dynamical systems},
  author={Kalman, Rudolf Emil},
  journal={Journal of the Society for Industrial and Applied Mathematics, Series A: Control},
  volume={1},
  number={2},
  pages={152--192},
  year={1963},
  publisher={SIAM}
}

@incollection{kitagawa1996linear,
  title={Linear {Gaussian} state space modeling},
  author={Kitagawa, Genshiro and Gersch, Will},
  booktitle={Smoothness priors analysis of time series},
  pages={55--65},
  year={1996},
  publisher={Springer}
}

@article{elvira2022graphical,
  title={Graphical inference in linear-{Gaussian} state-space models},
  author={Elvira, V{\'\i}ctor and Chouzenoux, Emilie},
  journal={IEEE Transactions on Signal Processing},
  volume={70},
  pages={4757--4771},
  year={2022},
  publisher={IEEE}
}

@article{rabiner1989tutorial,
  title={A tutorial on hidden {Markov} models and selected applications in speech recognition},
  author={Rabiner, Lawrence R},
  journal={Proceedings of the IEEE},
  volume={77},
  number={2},
  pages={257--286},
  year={2002},
  publisher={Ieee}
}

@book{cappe2005inference,
  title={Inference in hidden {Markov} models},
  author={Capp{\'e}, Olivier and Moulines, Eric and Ryd{\'e}n, Tobias},
  year={2005},
  publisher={Springer}
}

@article{ghojogh2019hidden,
  title={Hidden {Markov} model: tutorial},
  author={Ghojogh, Benyamin and Karray, Fakhri and Crowley, Mark},
  year={2019},
  journal={Engineering Archive}
}

@article{kalman1960new,
  title={A new approach to linear filtering and prediction problems},
  author={Kalman, Rudolph Emil},
  journal={Journal of Basic Engineering},
  volume={82},
  number={1},
  pages={35--45},
  year={1960}
}

@article{ghodsi2025many,
  title={How Many Heads Make an {SSM}? A Unified Framework for Attention and State Space Models},
  author={Ghodsi, Ali},
  journal={arXiv preprint arXiv:2512.15115},
  year={2025}
}

@book{simon2006optimal,
  title={Optimal state estimation: {Kalman}, {H} infinity, and nonlinear approaches},
  author={Simon, Dan},
  year={2006},
  publisher={John Wiley \& Sons}
}

@book{sarkka2023bayesian,
  title={Bayesian filtering and smoothing},
  author={S{\"a}rkk{\"a}, Simo and Svensson, Lennart},
  volume={17},
  year={2023},
  publisher={Cambridge university press}
}

@book{koller2009probabilistic,
  title={Probabilistic graphical models: principles and techniques},
  author={Koller, Daphne and Friedman, Nir},
  year={2009},
  publisher={MIT press}
}

@book{bishop2006pattern,
  title={Pattern recognition and machine learning},
  author={Bishop, Christopher M and Nasrabadi, Nasser M},
  volume={4},
  number={4},
  year={2006},
  publisher={Springer}
}

@article{ghojogh2024backpropagation,
  title={Backpropagation and Optimization in Deep Learning: Tutorial and Survey},
  author={Ghojogh, Benyamin and Ghodsi, Ali},
  journal={HAL preprints, hal-04694956},
  year={2024}
}

@article{rumelhart1986learning,
  title={Learning representations by back-propagating errors},
  author={Rumelhart, David E and Hinton, Geoffrey E and Williams, Ronald J},
  journal={nature},
  volume={323},
  number={6088},
  pages={533--536},
  year={1986},
  publisher={Nature Publishing Group}
}

@article{juang1991hidden,
  title={Hidden {Markov} models for speech recognition},
  author={Juang, Biing Hwang and Rabiner, Laurence R},
  journal={Technometrics},
  volume={33},
  number={3},
  pages={251--272},
  year={1991},
  publisher={Taylor \& Francis}
}

@article{mark2024application,
  title={The application of hidden {Markov} models in speech recognition},
  author={Mark, Gales and Steve, Young},
  journal={Foundations and Trends{\textregistered} in Signal Processing},
  volume={1},
  number={3},
  pages={195--304},
  year={2024},
  publisher={Emerald Publishing Limited}
}

@incollection{krogh1998introduction,
  title={An introduction to hidden {Markov} models for biological sequences},
  author={Krogh, Anders},
  booktitle={New comprehensive biochemistry},
  volume={32},
  pages={45--63},
  year={1998},
  publisher={Elsevier}
}

@article{yoon2009hidden,
  title={Hidden {Markov} models and their applications in biological sequence analysis},
  author={Yoon, Byung-Jun},
  journal={Current genomics},
  volume={10},
  number={6},
  pages={402--415},
  year={2009},
  publisher={Bentham Science Publishers}
}

@article{baum1966statistical,
  title={Statistical inference for probabilistic functions of finite state {Markov} chains},
  author={Baum, Leonard E and Petrie, Ted},
  journal={The annals of mathematical statistics},
  volume={37},
  number={6},
  pages={1554--1563},
  year={1966},
  publisher={JSTOR}
}

@article{baum1970maximization,
  title={A maximization technique occurring in the statistical analysis of probabilistic functions of {Markov} chains},
  author={Baum, Leonard E and Petrie, Ted and Soules, George and Weiss, Norman},
  journal={The annals of mathematical statistics},
  volume={41},
  number={1},
  pages={164--171},
  year={1970},
  publisher={JSTOR}
}

@article{kschischang2002factor,
  title={Factor graphs and the sum-product algorithm},
  author={Kschischang, Frank R and Frey, Brendan J and Loeliger, H-A},
  journal={IEEE Transactions on information theory},
  volume={47},
  number={2},
  pages={498--519},
  year={2002},
  publisher={IEEE}
}

@article{sepanj2025uncertainty,
  title={Uncertainty-Aware $\delta$-{GLMB} Filtering for Multi-Target Tracking},
  author={Sepanj, M Hadi and Moradi, Saed and Azimifar, Zohreh and Fieguth, Paul},
  journal={Big Data and Cognitive Computing},
  volume={9},
  number={4},
  pages={84},
  year={2025},
  publisher={MDPI}
}

@article{nazemi2024particle,
  title={Particle-Filtering-based Latent Diffusion for Inverse Problems},
  author={Nazemi, Amir and Sepanj, Mohammad Hadi and Pellegrino, Nicholas and Czarnecki, Chris and Fieguth, Paul},
  journal={arXiv preprint arXiv:2408.13868},
  year={2024}
}

@article{roesser2003discrete,
  title={A discrete state-space model for linear image processing},
  author={Roesser, Robert},
  journal={IEEE transactions on automatic control},
  volume={20},
  number={1},
  pages={1--10},
  year={2003},
  publisher={IEEE}
}
\bibliographystyle{icml2025}

\end{document}